\documentclass[10pt,twocolumn,letterpaper]{article}

\usepackage{cvpr}
\usepackage{times}
\usepackage{epsfig}
\usepackage{graphicx}
\usepackage{amsmath}
\usepackage{amssymb}
\usepackage{algorithm}
\usepackage{algorithmic}
\usepackage{diagbox}
\usepackage{float}
\usepackage{multirow}
\usepackage{multicol}

\usepackage[pagebackref=true,breaklinks=true,letterpaper=true,colorlinks,bookmarks=false]{hyperref}

\cvprfinalcopy 

\def\cvprPaperID{10099} 

\ifcvprfinal\fi
\begin{document}

\title{Clean-Label Backdoor Attacks on Video Recognition Models}

\author{Shihao Zhao\textsuperscript{1}\ \  
Xingjun Ma\textsuperscript{2}\footnotemark[2] \ \  
Xiang Zheng\textsuperscript{2} \ \
James Bailey\textsuperscript{2}\ \
Jingjing Chen \textsuperscript{1} \ \ Yu-Gang Jiang \textsuperscript{1}\footnotemark[2]\\
  \textsuperscript{1}Fudan University \ \ \ \ \ \textsuperscript{2}The University of Melbourne\\
}

\maketitle

\renewcommand{\thefootnote}{\fnsymbol{footnote}} 
\footnotetext[2]{Correspondence to: Xingjun Ma (xingjun.ma@unimelb.edu.au) and Yu-Gang Jiang (ygj@fudan.edu.cn)} 

\begin{abstract} 
\label{abstract}

Deep neural networks (DNNs) are vulnerable to backdoor attacks which can hide backdoor triggers in DNNs by poisoning training data. A backdoored model behaves normally on clean test images, yet consistently predicts a particular target class for any test examples that contain the trigger pattern. As such, backdoor attacks are hard to detect, and have raised severe security concerns in real-world applications.
Thus far, backdoor research has mostly been conducted in the image domain with image classification models. 
In this paper, we show that existing image backdoor attacks are far less effective on videos, and outline 4 strict conditions where existing attacks are likely to fail: 1) scenarios with more input dimensions (eg. videos), 2) scenarios with high resolution, 3) scenarios with a large number of classes and few examples per class (a ``sparse dataset"), and 4) attacks with access to correct labels (eg. clean-label attacks). 
We propose the use of a universal adversarial trigger as the backdoor trigger to attack video recognition models, a situation where backdoor attacks are likely to be challenged by the above 4 strict conditions.
We show on benchmark video datasets that our proposed backdoor attack can manipulate state-of-the-art video models with high success rates by poisoning only a small proportion of training data (without changing the labels).
We also show that our proposed backdoor attack is resistant to state-of-the-art backdoor defense/detection methods, and can even be applied to improve image backdoor attacks. 
Our proposed video backdoor attack not only serves as a strong baseline for improving the robustness of video models, but also provides a new perspective for more understanding more powerful backdoor attacks.(\href{https://github.com/ShihaoZhaoZSH/Video-Backdoor-Attack}{https://github.com/ShihaoZhaoZSH/Video-Backdoor-Attack})
\end{abstract}


\section{Introduction}
\label{Introduction}

\newcommand{\tabincell}[2]{\begin{tabular}{@{}#1@{}}#2\end{tabular}}
\begin{table*}[htbp]
\label{tab:1}
\centering
\setlength{\tabcolsep}{1mm}{
\begin{tabular}[htbp]{c|ccccc|c}
\hline
 & CIFAR-10  & CIFAR-100 & VOC2012-10 & UCF-10 & UCF-101 & UCF-101 (ours)\\
\hline
Success rate (\%) & 78.2  & 43.7 & 25.2 & 47.2 & 1.1 & \textbf{82.2}\\
\hline
Sparse dataset & $\times$ & $\surd$ & $\times$ & $\times$ & $\surd$ & $\surd$\\
\hline
High resolution & $\times$ & $\times$ & $\surd$ & $\times$ & $\surd$ & $\surd$\\
\hline
Video & $\times$ & $\times$ & $\times$ & $\surd$ & $\surd$ & $\surd$\\
\hline
\end{tabular}
}
\caption{ The attack success rate (\%) of existing clean-label backdoor attack Turner~\etal~\cite{DP:turner2019cleanlabel} under different strict conditions. The attack was applied to poison 30\% of training examples in the target class using trigger size of 1\% image area. "High resolution" refers to images/frames of size 224 $\times$ 224 (compared with size 32 $\times$ 32), while "sparse dataset" refers to CIFAR-100 \cite{krizhevsky2009learning} and UCF-101 \cite{DBLP:journals/corr/abs-1212-0402} (compared with other 10-class datasets). Datasets VOC2012-10 and UCF-10 consist of 10 random classes from VOC2012 \cite{pascal-voc-2012} and UCF-101 \cite{DBLP:journals/corr/abs-1212-0402} respectively. We use target model ResNet50~\cite{DLA:he2016resnet} for all image datasets and I3D~\cite{DBLP:journals/corr/CarreiraZ17} for video datasets and the first class (alphabetical order) as the target class. The last column shows result of our proposed attack under the most strict condition (eg. UCF-101). }
\label{tab:1_a}
\end{table*}

Deep neural networks (DNNs) are a family of powerful models that have been widely used to achieve state-of-the-art performance in many applications such as image classification \cite{DLA:he2016resnet}, natural language processing \cite{DLA:sutskever2014sequence_translation} and video recognition \cite{DBLP:journals/corr/CarreiraZ17}. 
Despite great success, DNNs have been criticized due to their low transparency, poor interpretability, and more importantly vulnerabilities to adversarial attacks \cite{AT:szegedy2013intriguing,Goodfellow2014ExplainingAH,ma2018characterizing,wang2019convergence,wu2020skip,Wang2020Improving} and backdoor attacks \cite{DP:gu2017badnets,DP:chen2017targeted,TROJAN,DP:turner2019cleanlabel,xie2019dba}. This has raised  concerns for the development of DNNs in applications such as face recognition \cite{sharif2016accessorize,dong2019efficient}, autonomous driving \cite{evtimov2017robust}, video analysis \cite{jiang2019black}, and medical diagnosis \cite{finlayson2019adversarial,ma2019understanding}.


\begin{figure}[t]
\begin{center}
\includegraphics[width=0.9\linewidth]{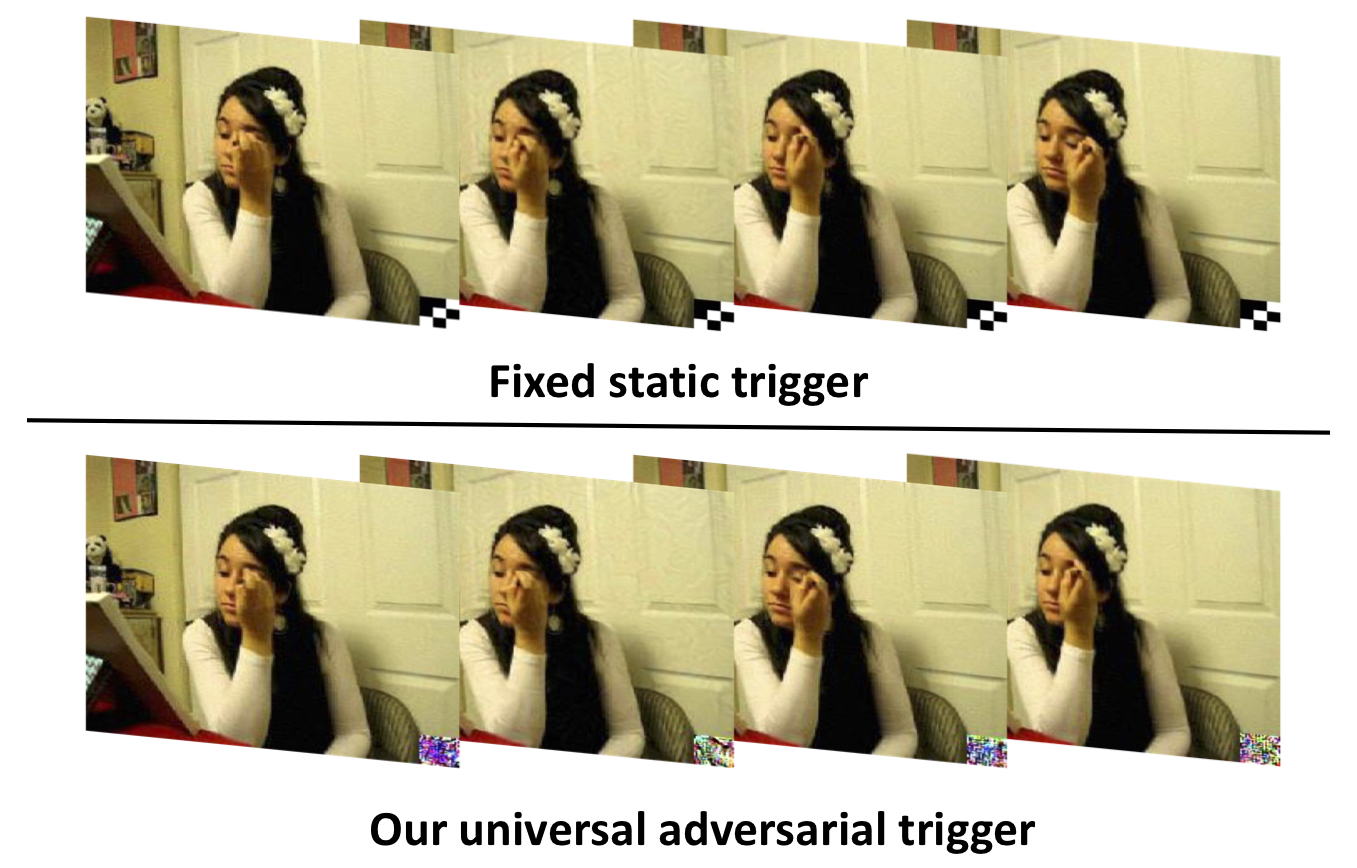}
\end{center}
   \caption{An example of clean-label poisoned videos by clean-label backdoor attack Turner~\etal~\cite{DP:turner2019cleanlabel} (top) and our proposed attack with universal adversarial backdoor trigger (bottom).}
\label{fig:1_a}
\end{figure}


Compared with adversarial attacks which are test time attacks but on  clean models, backdoor attacks pose more severe  threats by installing a hidden trigger into a target model at training time.
In particular, a backdoor attack poisons training data with a backdoor trigger (or pattern) so as to set up a link between the backdoor pattern and a target label. Models trained on poisoned data will remember the trigger pattern, and at test time, consistently predict a particular target class whenever the trigger pattern is present. Backdoor attacks are hard to detect since backdoored models still perform accurately on clean validation or test data.
Backdoor attacks can happen in scenarios when users download DNN models from an untrusted party or train models on data collected from unreliable sources, which is quite common in deep learning.
The study of backdoor attacks has thus become crucial for secure deep learning.

Existing backdoor attacks can be categorized into two types: 1) poison-label attacks which not only poison training examples but also change training labels (to the target class); and 2) clean-label attacks which poison only training examples while leaving training labels unchanged. Compared with poison-label attacks, clean-label attacks are more stealthy and resistant to data filtering or detection techniques \cite{DP:turner2019cleanlabel}.
While existing clean-label backdoor attacks have mostly been studied in the image domain with image classification models, their effectiveness in more demanding conditions such as on videos is still unexplored. 
In this paper, we close this gap by proposing the first video backdoor attack. 

Surprisingly, we find that existing backdoor attacks are far less effective on videos, and outline 4 strict conditions where existing image backdoor attacks are likely to fail: 1) scenarios with more input dimensions (eg. videos vs images), 2) scenarios with high resolution (eg. 224 $\times$ 224 vs 32 $\times$ 32), 3) scenarios with a large number of classes having very few examples per class (a ``sparse dataset"), and 4) attacks with access to the correct labels (eg. clean-label attacks). 
We use 5 different datasets to simulate the first 3 strict conditions under the clean-label setting (eg. the fourth condition), and test the attack success rate of one state-of-the-art clean-label backdoor attack Turner~\etal~\cite{DP:turner2019cleanlabel} under different strict conditions in Table \ref{tab:1}. 
We find that the existing attack Turner~\etal~\cite{DP:turner2019cleanlabel} has low effectiveness under certain conditions, and fails almost completely (attack success rate is only $1.1\%$) on video dataset UCF-101 where all the 4 strict conditions are satisfied.
These results emphasize the necessity of designing effective clean-label backdoor attack under more realistic conditions.


To address the limitations of existing backdoor attacks under strict conditions, in this paper, we propose a universal adversarial trigger for backdoor attack against video recognition models. 
The universal adversarial trigger is generated by exploiting adversarial technique \cite{Goodfellow2014ExplainingAH,AT:szegedy2013intriguing} to minimize the classification loss from non-target classes to the target class. When used to poison training data, we also apply adversarial perturbation on the target image to force the target model to focus more on the trigger pattern. Different to universal adversarial perturbation \cite{metzen2017universal,moosavi2017universal} crafted to fool DNN models at test time (eg. adversarial attacks), here we exploit it as a trigger pattern for more powerful backdoor attacks.
As shown in Table \ref{tab:1}, using the same poisoning rate and trigger size, our proposed attack can achieve a significantly higher success rate of 82.2\%. An example of our generated universal adversarial trigger is illustrated in Figure \ref{fig:1_a}.

Our main contributions are:
\begin{itemize}
 \item We study the problem of clean-label backdoor attack against video recognition models, and develop a novel approach to execute backdoor attacks under strict conditions. To the best of our knowledge, this is the first work on video backdoor attacks.
 
 \item We propose a universal adversarial trigger and two types of adversarial perturbations for more effective backdoor attack on videos, and demonstrate, on two benchmark video datasets against two state-of-the-art video recognition models, that our proposed attack is far more effective than existing backdoor attacks.
 
 \item We show that our video backdoor attack: 1) cannot be completely avoided by existing backdoor detection methods; 2) can generally be applied to improve backdoor attacks on image models.
\end{itemize}

\section{Related work}
\label{Related work}
We review existing backdoor attacks proposed for image classification models, backdoor defense methods, and state-of-the-art DNN models used for video recognition.


\subsection{Backdoor Attack}
\label{Backdoor Attacks}
Backdoor attack is a type of data poisoning attack that injects some trigger pattern into the training data so that the trained model will make incorrect predictions whenever the trigger pattern is present.
Existing backdoor attacks can be categorized into two types: 1) poison-label attacks that not only poison training examples but also change their labels (to a target class); and 2) clean-label attacks that only poison training examples while leaving their labels unchanged.

\noindent\textbf{BadNets.}
\label{BadNets}
Gu~\etal~\cite{DP:gu2017badnets} first investigated backdoor attacks in the deep learning pipeline, and proposed the Badnets attack.
BadNets injects a trigger pattern (sticker or checkerboard) to a set of randomly selected training images. Given a target class of the attacker's interest, the poisoned images are usually selected from the other classes than the target class, and their labels will be changed to the target class. This is to explicitly associate the poisoned images to the target class. Different trigger patterns have been proposed for poison-label attacks. For example, an additional image attached onto or blended into the target image ~\cite{DP:chen2017targeted}, a fixed watermark on the target image \cite{DP:steinhardt2017certified_data_poisoning}, or one fixed pixel (for low resolution (32 $\times$ 32) images).
Since the poisoned images are mislabeled, poison-label attacks can be easily detected by simple data filtering or human inspection \cite{DP:turner2019cleanlabel}.

\noindent\textbf{Clean-label backdoor attacks.}
\label{Clean-label Backdoor Attacks}
To make the attack less obvious, Turner \etal \cite{DP:turner2019cleanlabel} proposed the clean-label backdoor attack which does not need to change the labels of poisoned images. Since the clean-label poisoned images still have labels that are consistent with their main contents, clean-label backdoor attacks are more difficult to detect. However, this significantly increases the difficulty of the attack as the trigger pattern is no longer strongly associated with the target class, thus may get filtered out easily as irrelevant information by convolutional operations.  Turner \etal \cite{DP:turner2019cleanlabel} further introduced two techniques to make the attack more effective: 1) latent space interpolation using GANs and 2) adversarial perturbations bounded by $\ell_p$-norm. Both techniques can force the model to learn the trigger pattern instead of the original contents of the image. Although high success rate can be achieved under easy conditions, i.e., on low-resolution image datasets, this attack has failed under strict conditions imposed by video datasets, as we have shown in Table \ref{tab:1}. In this paper, we propose a more powerful backdoor trigger to address the limitations of Turner \etal \cite{DP:turner2019cleanlabel} in attacking video recognition models.

\subsection{Backdoor Defense}
\label{Backdoor Attack Detection}
Several detection methods have been proposed to defend against backdoor attacks. The simplest and most natural way is to perform data augmentation by introducing flips and crops to eliminate the trigger pattern. Other detection methods exploit feature characteristics of backdoor images to train effective detection classifiers.

\noindent\textbf{Spectral signatures.}
\label{Spectral Signatures}
Tran \etal\cite{DP:tran2018spectral_backdoor} propose a method to avoid backdoor attacks by detecting the presence of poisoned examples in the training set, based on Spectral Signatures. The intuition is that poisoned data may appear abnormal in the latent DNN space compared with clean data in the same class. The user can remove the backdoor outliers from the training set via singular value decomposition (SVD), then retrain the model on the purified data.

\noindent\textbf{Neural cleanse.}
\label{Neural Cleans}
Wang \etal \cite{DP:wang2019neural_cleanse} propose Neural Cleanse to avoid backdoor attacks by examining whether or not a trained model is infected. Neural Cleanse exploits gradient information to reverse engineer possible triggers, then detect outliers (eg. triggers) using a robust statistics measure called median absolute deviation (MAD). Neural Cleanse is based on the assumption that smaller modifications are required to cause misclassification in an infected model compared with a clean model.


\subsection{Video Recognition Models}
\label{Video Recognition Models}
State-of-the-art video recognition models include the Inflated 3D ConvNet (I3D) \cite{DBLP:journals/corr/CarreiraZ17} and the CNN+LSTM \cite{2015arXiv150308909Y,2015arXiv150207209J}.
The I3D model is based on 2D ConvNet inflation and pooling kernels of traditional 2D CNNs and is able to learn hierarchical spatial-temporal information directly from videos. The CNN+LSTM model combines the advantages of both CNNs and LSTMs, that is, it utilizes CNN to extract spatial representations and LSTM to exploit the temporal information contained in successive frames. In addition, optical flow information is widely used in two-stream video recognition networks to improve the models' performance \cite{feichtenhofer2016convolutional}. In this paper, we will apply our attack to invade I3D and CNN+LSTM models while discussing the effect of optical flow information in backdoor attacks.

\section{Proposed Video Backdoor Attack}
\label{Proposed Video Backdoor Attack}
In this section, we introduce our proposed backdoor attack on video recognition models. We first describe our threat model and overview the attack pipeline, then outline how to generate and apply the proposed universal adversarial trigger for the attack.

\begin{figure*}[t]
\begin{center}
\includegraphics[width=0.9\linewidth]{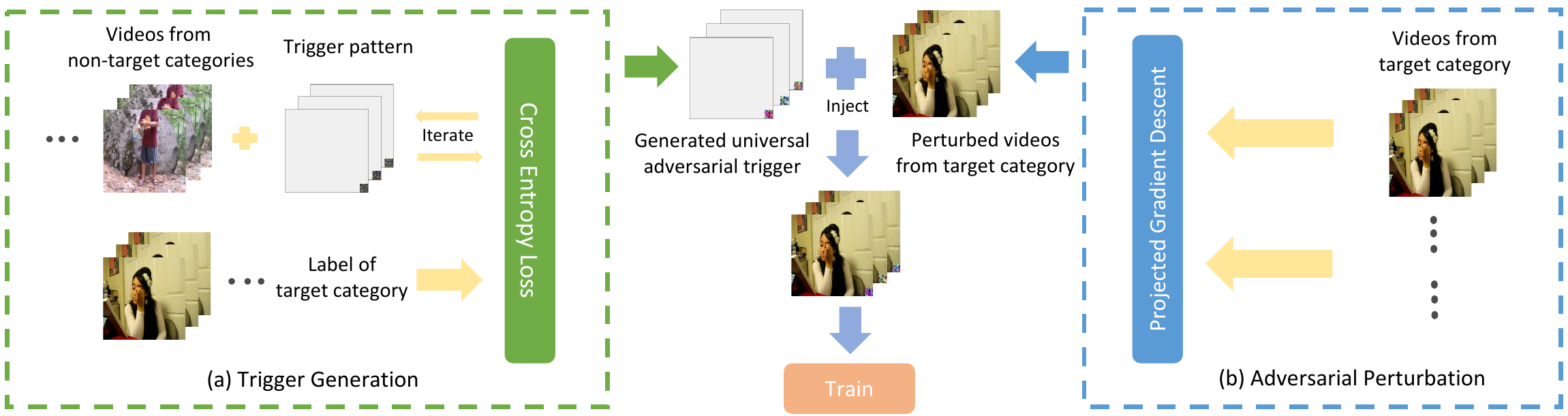}
\end{center}
   \caption{Overview of our attack pipeline. (a) Trigger Generation generates a universal adversarial trigger pattern specific to a task. (b) Adversarial Perturbation produces videos with manipulated features. We implement the attack by injecting the trigger into perturbed videos and providing them to users for training. We apply the same trigger pattern to videos during the test procedure.}
\label{fig:3_a}
\end{figure*}

\subsection{Threat Model}
\label{Threat Model}
The adversary in our setting is allowed to inject a small number of perturbed samples into the training set. The goal of the attack is to induce the network trained by end users to consistently predict a target class when the backdoor trigger is presented, but behave normally on clean data. Specifically, the adversary is assumed to know the
architecture of the network that the end user uses
and have access to training data, but has no permission to 
know any other 
configuration of the user's training procedure. 
For stealthiness, we implement our attack under the clean-label setting.

\subsection{Overview of the Attack Pipeline}
\label{Overview of the Attack Pipeline}
The structure of our proposed pipeline is illustrated in Figure \ref{fig:3_a}. It consists of three steps: (a) Trigger generation. Given a clean training set and a
clean-trained
model on the data, this step will generate a universal adversarial trigger using gradient information through iterative optimization. (b) Adversarial perturbation. We minimize an adversarial loss using Projected Gradient Descent \cite{madry2017towards} (PGD) to produce videos with adversarial perturbations towards the target class. (c) Poisoning and inference. We inject the generated trigger (by step (a)) into perturbed videos (by step (b)) as poisoned samples for training. At the inference stage, we trick the target model trained on the poisoned data to predict the target class by attaching our universal adversarial trigger to a test video.


\subsection{Backdoor Trigger Generation}
\label{Backdoor Trigger Generation}


Given a clean-trained model on the clean training data, we optimize to find a trigger pattern that minimizes the cross entropy loss towards the target class. The trigger pattern is patched and optimized on videos from all non-target classes but relabeled to the target class.
This forces the network to predict the target class when the trigger pattern presents.

Specifically, given a trigger mask $m$, a trigger pattern $t$, videos in non-target categories $x$, one-hot vector of target label $y$ with dimension $l$, we generate a universal adversarial trigger pattern by minimizing the cross-entropy loss as following,
\begin{eqnarray}
    & \min\limits_{t} \sum_{i=1}^{M}-\frac{1}{l}\sum_{j=1}^{l}y_j \log(h_j(\tilde x_i)),
    \label{eq:1}
\end{eqnarray}
where $M$ is the total number of training samples from non-target classes, $h = F(\tilde x)$ is the softmax output of the clean-trained model, $\tilde x = (1-m) * x + m * t$ is the poisoned samples. By minimizing the above loss, we can find the universal (the same trigger pattern $t$ for all non-target videos) adversarial trigger for a certain training dataset targeted to a target class.


\begin{algorithm}
\caption{Universal Adversarial Trigger Generation}
\label{alg:1}
\begin{algorithmic}[1]
\REQUIRE ~~\\
Model $F$,Trigger Mask $m$, Learning Rate $\alpha$, 
Non Target Videos Set $S = \{(x^{(j)},y^{(j)})\}_{j=1}^{M}$, Target Label $\boldsymbol y$, Total Steps $N$, Trigger Size $w$, Batch Size $B$
\ENSURE Universal Trigger $t$
\STATE $t =  \mathrm{InitializeTrigger}(w)$
\FOR {$i$ in $\mathrm{range}(N)$}
\STATE $S_i = \{(x^{(j)},y^{(j)})\}_{j=1}^{B} = \mathrm{RandomlyPick}(S, B)$
\STATE $\tilde{x}^{(j)} = (1-m) * x^{(j)} + m * t, (x^{(j)},y^{(j)})\in S_i$
\STATE $h = F(\tilde{x}^{(j)})$
\STATE $L = \sum_{j=1}^{B}-\frac{1}{l}\sum_{k=1}^{l}[y^{(j)}_k \log(h_k)]$
\STATE $\delta=\frac{\partial{L}}{\partial{t}}$
\STATE $t = t - \alpha * \mathrm{sign}(\delta)$
\ENDFOR
\STATE return t
\end{algorithmic}
\end{algorithm}{}

The trigger pattern $t$ is generated as follows. 
We first take a small region at the bottom right corner of the input video as the trigger area, and randomly initialize the area while mask off the other areas. During training, we iteratively perturb the trigger area on different videos from all non-target classes (which are re-labeled to the target class). In other words, the perturbed trigger area on one video is kept and passed along to other videos for further perturbation.
The complete generation algorithm is described in Algorithm \ref{alg:1}.

\subsection{Enhancing Backdoor Trigger}
\label{Enhancing Backdoor Trigger}

In order to  enhance the backdoor trigger and make the features of the original videos less salient, we perform adversarial perturbation, a kind of unnoticeable and structured noise, to the clean video sample before applying our backdoor trigger.
This reduces the interference of the original content and encourage the network to pay more attention to the trigger \cite{DP:turner2019cleanlabel}.

Specifically, we randomly sample a certain proportion of videos in the target class and adversarially perturb them by PGD. These perturbed videos are then patched by our universal adversarial backdoor trigger (generated in the previous step) as poisoned samples. Formally, given a clean-trained model $F$ and a target video $x$, we construct the adversarial perturbation $\eta$ by maximizing the loss $L$ as,
\begin{equation}
    \begin{split}
    \max\limits_{\left\| \eta \right\|_\infty<\epsilon} L(x+\eta),
    \end{split}
    \label{eq:2}
\end{equation}
where, $\left\| \cdot \right\|_{\infty}$ is the $\ell_{\infty}$-norm, and $\epsilon$ is the maximum perturbation.
We introduce two types of perturbations (corresponding to two different loss functions). They are both intuitively useful for effective backdoor attacks and are empirically evaluated in Section \ref{Ablation Study}.

\noindent\textbf{Uniform adversarial perturbation.}
\label{Uniform Adversarial Perturbation}
It perturbs the output probability to a uniform distribution so as to weaken the learning of the original features towards any classes.
Accordingly, the loss function $L$ is,
\begin{eqnarray}
    L = \frac{1}{l}\sum_{j=1}^{l}\hat{y}_j\log(h_j),
    \label{eq:4}
\end{eqnarray}
where $h = F(\hat x)$ is the softmax output, $\hat x= x + \eta$ is the perturbed sample, $\hat{y}=[\frac{1}{l},\cdots,\frac{1}{l}]$, and $l$ is the number of classes. Note that maximizing this loss function is equivalent to minimizing the cross entropy loss with respect to $\hat{y}$. Videos with this uniform adversarial perturbation tend to lose any strong natural features. Thus, the model will more responsive to salient characteristics such as the backdoor trigger pattern.

\noindent\textbf{Targeted adversarial perturbation.}
\label{Targeted Adversarial Perturbation}
This perturbation has been proposed in the field of adversarial research for targeted adversarial attacks \cite{Goodfellow2014ExplainingAH,madry2017towards,AT:szegedy2013intriguing}.
Given an input video $x$ from the target class $y$, the loss function $L$ is defined as,
\begin{eqnarray}
    L = -\frac{1}{l}\sum_{j=1}^{l}y_j \log(h_j),
    \label{eq:3}
\end{eqnarray}
where $h = F(\hat x)$ is the softmax output, $\hat x= x + \eta$ is the perturbed sample and $l$ is the number of classes. Empirically, we find that targeted perturbations may switch the network's output from one class to another with overconfident prediction. Thus, when applied with these perturbations, the perturbed video tends to have a strong pattern towards another class in deep feature space. This also forces the target model to capture the trigger pattern during training.

\subsection{Attack with the Universal Adversarial Trigger}
\label{Attack with the Backdoor Trigger}
\noindent\textbf{Poisoning and inference.}
\label{Poisoning and Inference}
To complete our attack, we inject the universal adversarial trigger generated in Section \ref{Backdoor Trigger Generation} into perturbed videos obtained in Section \ref{Enhancing Backdoor Trigger}. The poisoned videos are provided to users along with the rest of clean videos for training. At the inference stage, we patch the same trigger to test videos and the infected model will be manipulated to output the target class. At this point, we have implemented a complete backdoor attack.

\noindent\textbf{Analysis.}
\label{Analysis}
We briefly emphasize here that we study clean-label backdoor attacks in video from two complementary aspects. The universal adversarial trigger is the foundation of our method. It contains abundant inherent information which caters to the prior distribution of the target class, which makes video attacks possible and practical. Two different adversarial perturbations significantly interfere with the original features of videos in the target class to induce the model to learn more about the trigger pattern.

\section{Experiments}
\label{Experiments}

\begin{figure*}[t]
\begin{center}
\includegraphics[width=0.9\linewidth]{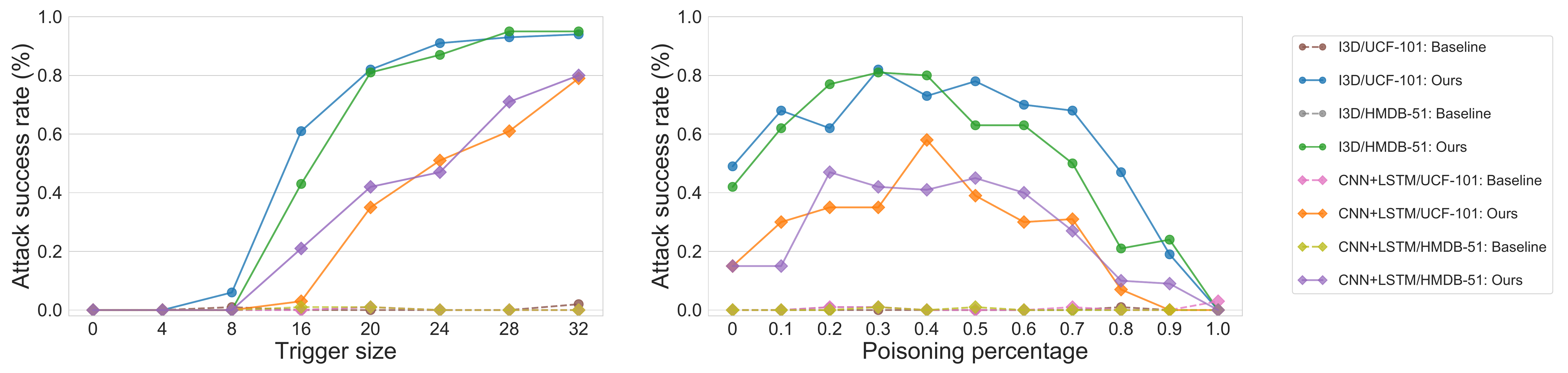}
\end{center}
   \caption{The attack success rates with respect to two factors: trigger size (left) and poisoning percentage (right). We set poisoning percentage to 30\% (which accounts for 0.3\%/0.6\% of all the data in UCF-101/HMDB-51) when we vary the trigger size, while setting trigger size to be $20 \times 20$ when varying the poisoning percentage. For all experiments, we choose the first class (in alphabetical order) to be the target label (eg. “ApplyEyeMakeup” in UCF-101 and “brush hair” in HMDB-51)}
\label{fig:4_2b}
\end{figure*}

\begin{table*}[]
    \centering
    \setlength{\tabcolsep}{2mm}{
    \begin{tabular}{ccccccccccc}
    \hline
     Method $\setminus$ class & \tabincell{c}{Apply \\ EyeMakeup}& Biking & \tabincell{c}{Clean\&\\ Jerk}& \tabincell{c}{Frisbee \\ Catch} & \tabincell{c}{Horse \\ Race} & \tabincell{c}{Long \\ Jump} & \tabincell{c}{Playing \\ Dhol} & Punch & Skiing & Taichi\\
    \hline
    Baseline & 1.1 & 0.5 & 0.0 & 3.8 & 0.1 & 0.0 & 2.9 & 0.6 & 0.0 & 2.6\\
    \hline
    Ours & 82.2 & 76.2 & 87.5 & 88.0 & 70.2 & 74.9 & 91.3 & 82.5 & 81.7 & 86.0\\
    \hline
    \end{tabular}
    }
    \caption{Comparison of the attack success rates (\%) on 10 different categories of UCF-101 against I3D model.}
    \label{tab:4_2a}
\end{table*}


In this section, we evaluate the effectiveness of our proposed backdoor attack on video recognition models, and resistance to state-of-the-art backdoor detection methods. We also conduct a comprehensive ablation study on various aspects of our attack.

\subsection{Experimental Settings}\label{sec:experimental_setting}
\label{Experimental Settings}

\noindent\textbf{Datasets and DNN models.}
\label{Datasets and DNN models}
We consider two benchmark datasets for video recognition: UCF-101 \cite{DBLP:journals/corr/abs-1212-0402} and HMDB-51 \cite{HMDB51}, and two state-of-the-art video recognition models: I3D and  CNN+LSTM. For I3D, we use a kinetics-400 pretrained model to initialize and sample 64 frames per video for finetuning on UCF-101 and HMDB-51. For CNN+LSTM, we use a fixed ImageNet pretrained ResNet50 as the CNN feature extractor and train LSTM on top of it. Input video frames are subsampled by keeping one out of every 5 for CNN+LSTM model. The test accuracy of these models can be found in Table \ref{tab:4_1a}. We use these clean-trained models to generate the universal adversarial triggers and adversarial perturbations. The size of input frame in both two models is set to 224 $\times$ 224. Here, we only consider RGB models, and later in Section \ref{Attacking Two-stream Video Models}, we will analyze our attacks on two-stream networks that consist of RGB and optical flow information.

\begin{table}[H]
\begin{center}
\setlength{\tabcolsep}{3mm}{
\begin{tabular}{ccc}
\hline
Model & UCF-101 & HMDB-51\\
\hline
I3D & 91.5 & 63.4\\
\hline
CNN+LSTM & 76.6 & 45.3\\
\hline
\end{tabular}
}
\end{center}
\caption{Test accuracy (\%) of the video models.}
\label{tab:4_1a}
\end{table}

\begin{figure}[t]
\begin{center}
\includegraphics[width=0.90\linewidth]{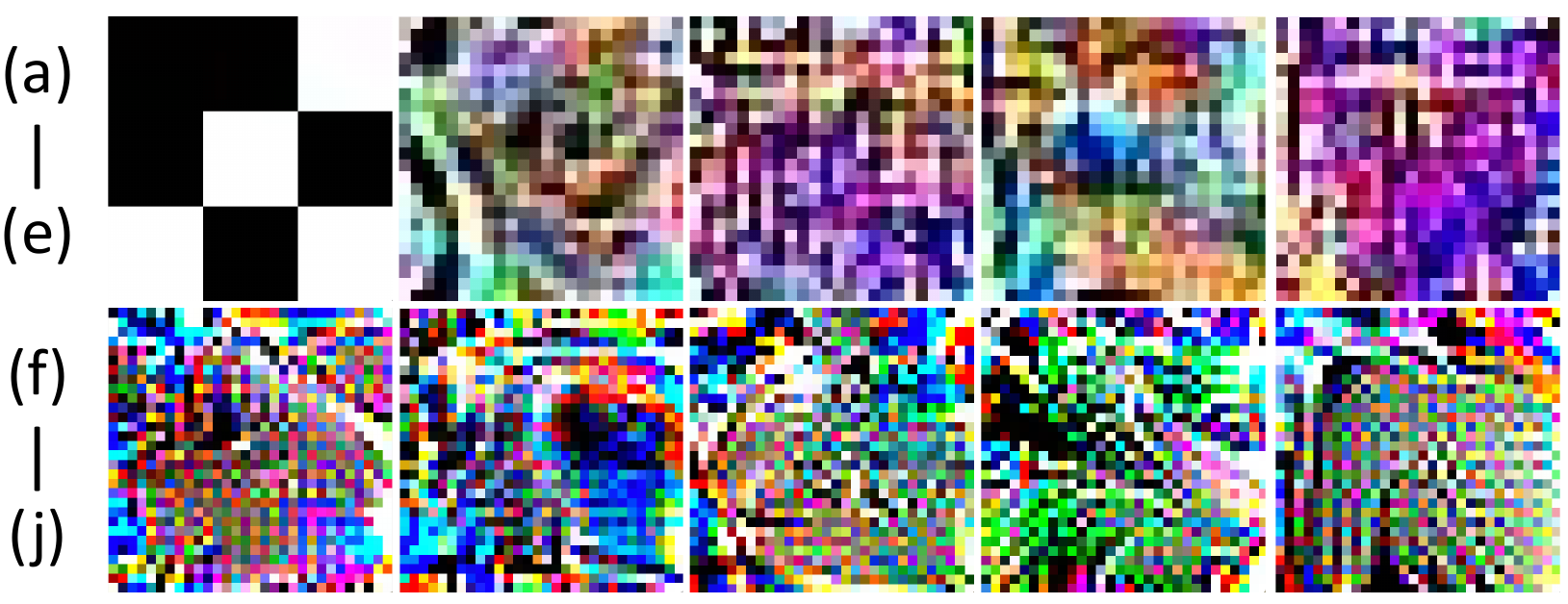}
\end{center}
   \caption{Examples of different triggers: (a) a fixed static trigger we use in both video and image tasks as our baseline. (b)-(e) our universal adversarial trigger on 4 different frames in a video for target class ``ApplyEyeMakeup” in UCF-101 against I3D model. (F)-(G): universal adversarial triggers for 5 categories (eg. ``aeroplane", ``bus”, ``diningtable”, ``pottedplant” and ``tvmonitor”) from VOC2012 image classification.}
\label{fig:4_2a}
\end{figure}

\noindent\textbf{Baselines and attack setting.}
\label{Baseline and Attack setting}
We transfer the image-based clean-label backdoor attack in \cite{DP:turner2019cleanlabel} directly to video frames as our baseline. For the baseline attack, we choose the PGD method bounded in $\ell_\infty$-norm to apply adversarial perturbations (also for our targeted adversarial perturbations). Then, we install a fixed static trigger (Figure \ref{fig:4_2a}) into frames of poisoned videos. We implement our attack following the pipeline in Section \ref{Proposed Video Backdoor Attack} and utilize the targeted adversarial perturbations here for fair comparison. It is worth mentioning that the accuracy of the infected models on clean test set has no obvious decline compared with clean trained models (in some cases, the test accuracy are even higher).
All our attacks are applied under the clean-label setting. We generate our universal adversarial triggers using Algorithm \ref{alg:1}, with learning rate $\alpha=1$ for 2000 steps. In each step, we randomly sample 10 videos to calculate the average gradient. For adversarial perturbation, we optimize Eqn. \eqref{eq:2} using PGD with $\epsilon=16$.

\noindent\textbf{Evaluation metric.}
\label{Evaluation metric}
The attack success rate (ASR) is selected as the evaluation metric, which is the fraction of inputs not labeled as the target class but misclassified to the target class after the backdoor trigger is applied. All experiments are run on a GeForce GTX 1080Ti GPU.

\subsection{Effectiveness of Our Approach}
\label{Effective of Our Approach}

\noindent\textbf{Attack success rates with varying trigger sizes.}
\label{Attack success rates with varying trigger sizes}
We first evaluate the attack performance under different trigger sizes. 
The results are shown in the left subfigure of Figure \ref{fig:4_2b}. 
As can be observed, the baseline attack has an extremely low level of attack success rate even when the trigger size is as large as 32, while the attack success rate of our attack rises rapidly with the increase of the trigger size and reach a plateau eventually. Against I3D model on UCF-101, our proposed attack achieves 61.2\% even at a small trigger size of 16 (which only accounts for 0.005\% of the total image area). When the trigger size increases to 28, our attack can successfully attack the target model 93.7\% of the time. 
In general, larger trigger size leads to stronger attack. 
However, this will inevitably make the trigger more conspicuous. A trade-off should be made between attack success rate and stealthiness in real-world application.

\noindent\textbf{Attack success rates under different poisoning percentages.}
\label{Attack success rates under different poisoning percentages}
We then demonstrate the impact of poisoning percentage (eg. the proportion of poisoned videos in the target class) on the attack performance. We choose the same class used above as the target class and fix trigger size to $20 \times 20$. The attack success rate with respect to different poisoning percentages is shown in the right subfigure of Figure \ref{fig:4_2b}.
Besides the poor performance of the baseline attack, we find that attack success rate does not increase monotonically with the poisoning percentage, instead, it first rises then drops. The attack success rate is over $60\%$ when poisoning percentage varies from 20\% to 70\%, but decreases dramatically once out of this range (even reduces to nearly 0\% at poisoning percentage of 100\%). We suspect this surprising phenomenon is caused by the following reason. The universal adversarial trigger is designed to reflect the inherent pattern of the target class. If the poisoning percentage is excessively high that there are few clean videos are left in the target class, the model will learn almost nothing about the original features of the target class and turn to new features that are more salient than the trigger generated by the clean-trained model, resulting in a rapid decline in performance.
We further choose 10 categories in UCF-101 as the target class respectively against the I3D model under the best poisoning percentage of 30\% (trigger size is set to 20 $\times$ 20). The results are shown in Table \ref{tab:4_2a}. Our attack performs well and is significantly better than the baseline attack across all the 10 target categories.

\begin{table*}[htbp]
\centering
\setlength{\tabcolsep}{2mm}{
\begin{tabular}{c|ccc|ccc|ccc}
\hline
\multirow{2}*{} & \multicolumn{3}{c|}{Fixed static trigger} & \multicolumn{3}{c|}{Randomly sampled trigger} & \multicolumn{3}{c}{Universal adversarial trigger}\\
~ & None & targeted & uniform & None & targeted & uniform & None & targeted & uniform\\
\hline
UCF-101 & 0.0 & 0.0 & 0.0 & 2.2 & 0.0 & 1.7 & 61.1 & \textbf{82.2} & 76.9\\
\hline
HMDB-51 & 0.0 & 1.4 & 0.0 & 0.0 & 4.1 &0.0  & 69.2 & 81.0 &\textbf{81.7}\\
\hline
\end{tabular}
}
\caption{Attack success rates (\%) of our proposed attack with three different types of triggers: fixed static trigger, randomly dynamic trigger (which has different random patterns among frames), universal adversarial trigger, and two different perturbations: no perturbation, targeted perturbation and uniform perturbation. Best results are highlighted in \textbf{bold}.}
\label{tab:4_3a}
\end{table*}

\subsection{Ablation Study}
\label{Ablation Study}
To better understand our attack, we perform extensive ablation studies with 3 different types of triggers enhanced by two different adversarial perturbations (eg. uniform versus targeted). We train I3D models on UCF-101/HMDB-51 and choose “ApplyEyeMakeup” as the target category. We set trigger size to be 20 $\times$ 20 and poisoning percentage to be 30\%. The results are shown in Table \ref{tab:4_3a}.

\begin{figure}[h]
\begin{center}
\includegraphics[width=0.9\linewidth]{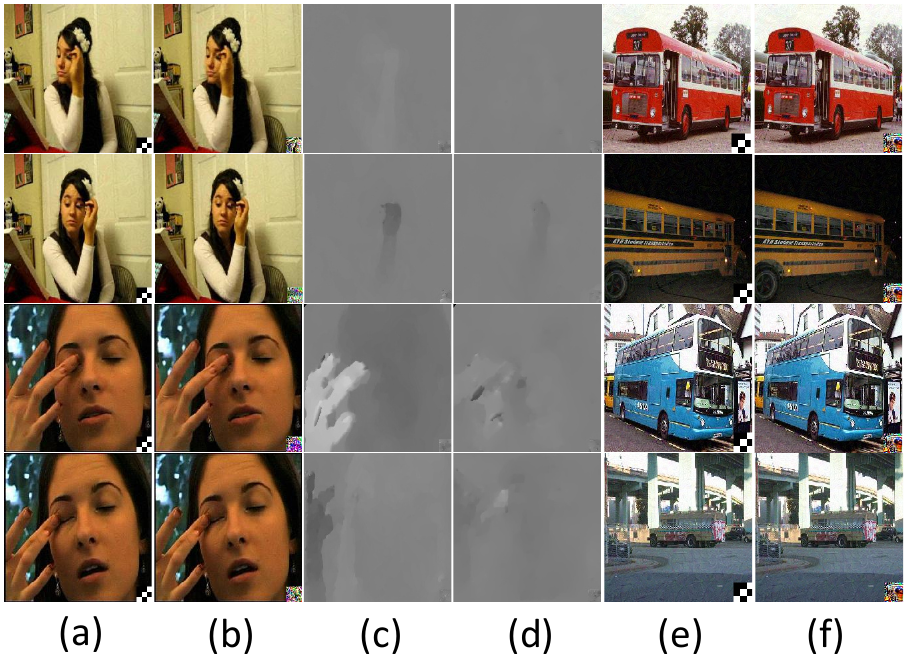}
\end{center}
   \caption {Examples of different poisoned data. (a)/(b): videos poisoned by fixed static trigger/universal adversarial trigger. (c)/(d): optical flow information of corresponding videos in (b) in x/y directions. (e)/(f): images poisoned by fixed static trigger/universal adversarial trigger (respectively for 4 target categories in 4 different experiments) .}
\label{fig:4_showall}
\end{figure}

\begin{figure}[t]
\begin{center}
\includegraphics[width=0.9\linewidth]{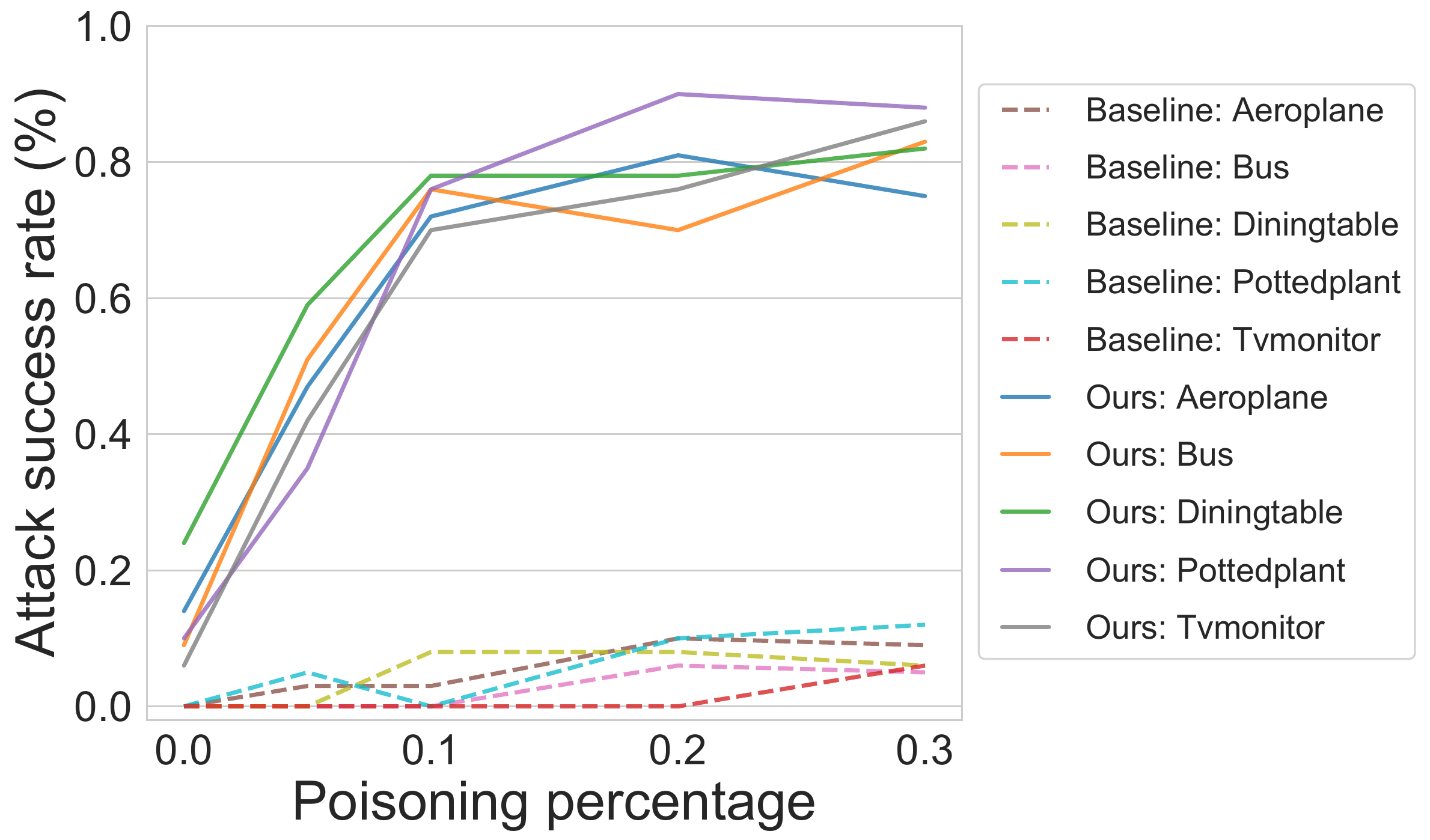}
\end{center}
   \caption{Comparative results on the image task. The target model is a ResNet50 with input size 224 $\times$ 224. We follow the same pipeline in \cite{DP:turner2019cleanlabel} as our baseline and apply the same targeted adversarial perturbation (for enhancement) for both attacks. The trigger size is 30 $\times$ 30 and $\epsilon$ in Eqn. \eqref{eq:2} is 8. (All triggers are illustrated in Figure \ref{fig:4_2b})}
\label{fig:4_5a}
\end{figure}

We first explore three different types of triggers: 1) fixed static trigger, 2) randomly sampled trigger, and 3) universal adversarial trigger. As can be seen from Table \ref{tab:4_3a}, neither the fixed static trigger nor the randomly sampled trigger are effective. The usage of universal adversarial trigger drastically increases the attack success rate. Both of the two different adversarial enhancements (eg. targeted versus uniform) can improve success rate by around 10\% - 20\%, with targeted adversarial perturbation more effective on UCF-101 dataset while uniform adversarial perturbation more effective on HMDB-51 dataset.


\subsection{Attacking Two-stream Video Models}
\label{Attacking Two-stream Video Models}
Optical flow information is often exploited to improve the performance of video recognition models. Here, we test this factor in our attack. We utilize both RGB and optical flow as inputs to construct a two-stream network. For our attack, we first inject the RGB trigger to videos, then generate optical flow of these poisoned videos using TVL1 algorithm \cite{ipol.2013.26}. We choose the average function to fuse the two streams and test on UCF-101 with I3D model. We visualize an example of poisoned RGB inputs and its corresponding optical flow input in Figure \ref{fig:4_showall}.

We find that the optical flow inhibits backdoor attack to some extent.
When trigger size is set to 20 $\times$ 20 and poisoning percentage 0.7, the attack success rate is 15.2\% on optical flow network, 68.5\% on RGB network and 54.7\% on fused two-stream network. This degradation of performance is mainly attributed to the independence of the RGB space to the optical flow space, which makes transfer of the universal adversarial trigger generated in the RGB space less effective in the optical flow space.



\subsection{Improving Image Backdoor Attacks}
\label{Improving Image Backdoor Attacks}
Here, we explore the generalization capability of our attack on images against image classification models. The experiments are conducted on VOC2012 which is a ``sparse dataset"(20 classes, around 400 images per class) and high resolution (224 $\times$ 224). We randomly choose 5 target categories and test the attack success rates under different poisoning percentages. The results are shown in Figure \ref{fig:4_5a}. Again, our method can effectively improve the attack performance under these strict conditions (eg. sparse dataset and high resolution) with images. The results confirm the effectiveness and generalization of our proposed attack on images, especially under strict conditions. 

\begin{table}[]
\vspace{-0.6cm}
\setlength{\tabcolsep}{1.5mm}{
\begin{tabular}{ccccc}
\hline
Conditions & \#Clean & \#Poisoned & \tabincell{c}{\#Clean \\removed} & \tabincell{c}{\#Poisoned \\ removed}\\
\hline
\tabincell{c}{Trigger \\ Perturbation} & 71 & 30 & 2 & 28\\
\hline
Trigger & 71 & 30 & 1 & 29\\
\hline
Perturbation & 71 & 30 & 18 & 12\\
\hline
\end{tabular}
}
\caption{Detection performance of spectral signatures \cite{DP:tran2018spectral_backdoor} against our attack on I3D model and UCF-101 dataset. We use set $\varepsilon$ to 1.5, trigger size to 20 $\times$ 20, poisoning percentage to 30\% and seelct target class ``ApplyEyeMakeup".}
\label{tab:4_6a}
\end{table}


\subsection{Resistance to Backdoor Defense Methods}
\label{Resistance to Backdoor Attack Defense}
\noindent\textbf{Resistance to data augmentation.}
Data augmentation is a common technique to diversify datasets, which includes randomly sampling, cropping, or rotating some frames in video recognition tasks. This process might reduce the performance of backdoor attacks by randomly removing or destroying the trigger patched to the poisoned videos. 
To test whether it is an effective way to avoid our attacks, we do experiments using I3D trained on UCF-101. We set trigger size to 20 $\times$ 20, and poisoning percentage to 30\%.  With data augmentation, the attack success rate can still reach 56.3\%. This is because our universal adversarial trigger is powerful enough that it achieves great attack results even with an extremely low poisoning percentage (68.1\% at poisoning percentage of 0.001\% with respect to the entire dataset). In practice, data augmentation can be effectively evaded by simply increasing the poisoning percentage.


\noindent\textbf{Resistance to spectral signature detection.}
\label{Resistance to spectral signature based detection}
As mentioned in Section \ref{Spectral Signatures}, Tran \etal \cite{DP:tran2018spectral_backdoor} proposes Spectral Signatures to detect backdoor attacks by filtering suspected samples in training set. We conduct experiments to test whether or not this defense method can detect our attack. The results are shown in Table \ref{tab:4_6a}. 
We find that most of the poisoned videos are removed by this method (28/30 for ``Trigger and Perturbation" and 29/30 for ``Trigger") once the universal adversarial trigger presents. Their success may be caused by huge separation of distributions between poisoned and clean videos in latent space by our attack.
However, as we have empirically shown in the right subfigure of Figure \ref{fig:4_2b} that our attack can still achieve high success rate $> 40\% $ even when only $<1\%$ of data is poisoned.

\begin{figure}[t]
\vspace{-0.8cm}
\begin{center}
\includegraphics[width=0.9\linewidth]{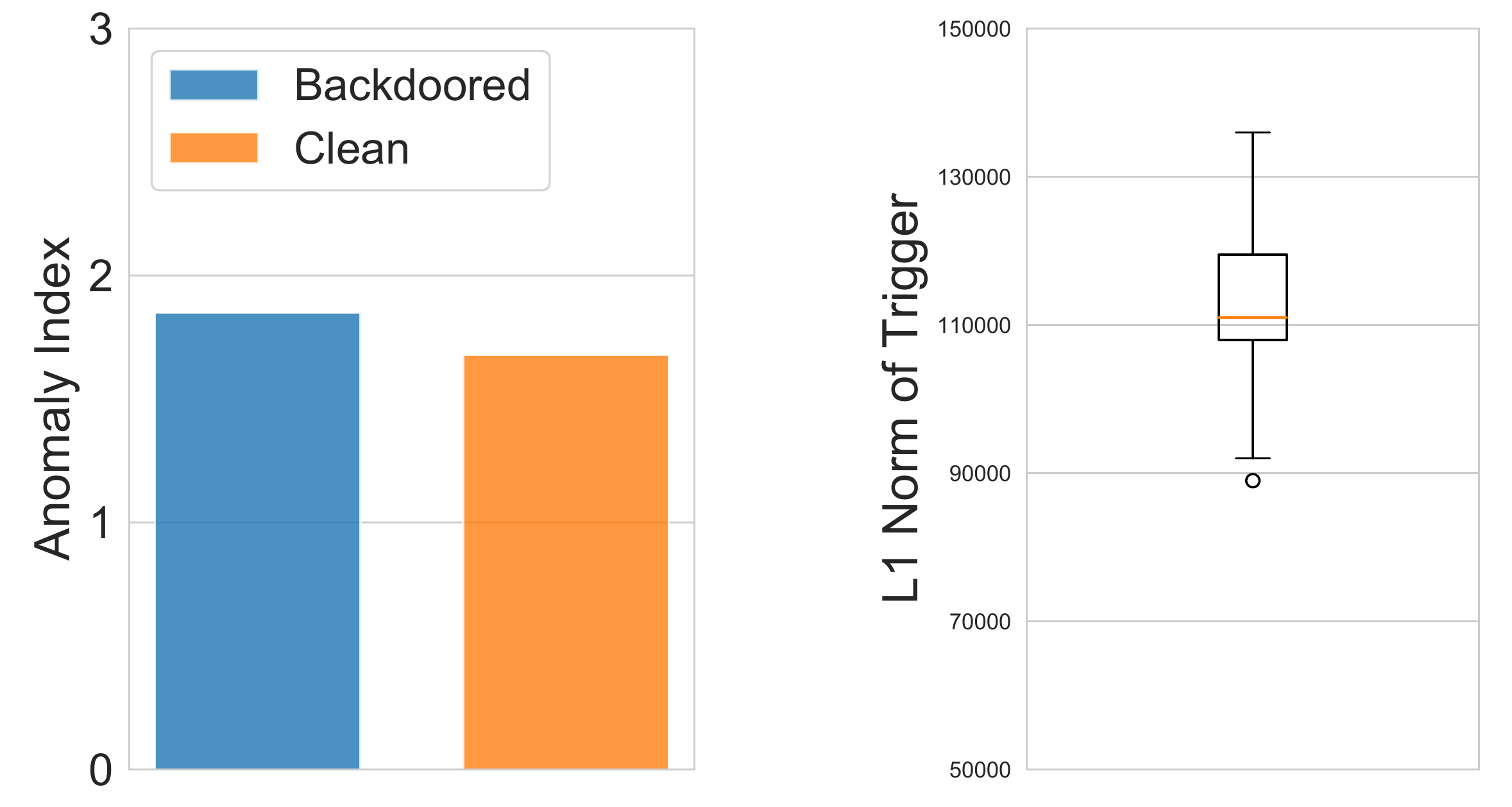}
\end{center}
   \caption{Backdoor using Neural Cleanse \cite{DP:wang2019neural_cleanse}. Left: Anomaly measurement of backdoored vs clean model by how much the class with the smallest trigger deviates from the remaining classes. An anomaly index $> 2$ indicates a detected backdoored model. Right: $\ell_1$-norm of triggers for infected vs uninfected classes in backdoored I3D model by our attack \cite{DP:wang2019neural_cleanse}. Box plot shows min/max and quarterlies, and the dot represents the target class. Detailed interpretation of the two plots can be find in \cite{DP:wang2019neural_cleanse}.}
\label{fig:4_6a}
\end{figure}

\noindent\textbf{Resistance to neural cleanse.}
\label{Resistance to neural cleanse}
As discussed in Section \ref{Neural Cleans}, Neural Cleans detects whether a trained model has been infected by backdoor attacks, for which it assumes that samples generally require smaller modifications to be misclassified into the target class. Here, we test the resistance of our proposed attack to Neural Cleans with I3D model on UCF-101 dataset, trigger size 20 $\times$ 20 and poisoning percentage 30\%. As shown in Figure \ref{fig:4_6a}, Neural Cleans fails to detect the backdoored I3D model by our attack, i.e., anomaly index $< 2$ for the backfoored model. This is because the modifications made by their reversed triggers has similar effect in the deep feature space as our universal adversarial trigger, thus no difference (or outlier) can be detected. Thereby, our proposed attack is resistant to Neural Cleanse.

\section{Conclusion}
\label{Conclusion}
In this paper, we have studied the problem of backdoor attack on video recognition models.
We outline 4 strict conditions posed by videos, and show existing backdoor attacks are likely to fail under these conditions. To address this, we propose the use of a universal adversarial trigger and two types of adversarial perturbation for more effective backdoor attacks with videos.
We show on benchmark video datasets that our proposed backdoor attack can manipulate state-of-the-art video models with high success rates by poisoning only a small proportion of training data.  We also show that our proposed backdoor attack is resistant to state-of-the-art backdoor detection methods to some extent, and can even be applied to  improve  image  backdoor  attacks.  Our proposed video backdoor attack can serve as a strong baseline for improving the robustness of video models.

\section*{Acknowledgement}
This work was supported in part by National Key Research and Development Program of China under Grant 2018YFB1004300.

\bibliography{egbib.bib}
\end{document}